\documentclass[final,5p,times,twocolumn]{elsarticle}

\usepackage{amssymb}
\usepackage{amsmath}
\usepackage{url}
\usepackage{soul}
\usepackage{xcolor}
\usepackage{mdframed}

\begin{document}

\begin{frontmatter}



\title{ Uncertainty-Aware Segmentation Quality Prediction via Deep Learning Bayesian Modeling: Comprehensive Evaluation and Interpretation on Skin Cancer and Liver Segmentation
}

%Other titles
% Uncertainty-Aware Semantic Segmentation: Bayesian Modeling and Quality Prediction Framework


\author[inst1]{Sikha O K}

\affiliation[inst1]{organization={BCN Medtech},%Department and Organization
            addressline={Department of Engineering,Universitat Pompeu Fabra},
            city={ Barcelona, Spain}
            }
\author[inst2]{Meritxell Riera-Marín}
\author[inst1]{Adrian Galdran}
\author[inst2]{Javier García López}
\author[inst2]{Júlia Rodríguez-Comas}
\author[inst1]{Gemma Piella}
\author[inst1,inst3]{Miguel A.\ González Ballester}
\affiliation[inst2]{organization={Scientific and Technical Department, Sycai Technologies S.L.},%Department and Organization
            addressline={ Barcelona, Spain}}
\affiliation[inst3]{organization={ ICREA},%Department and Organization
            addressline={ Barcelona, Spain}}

\begin{abstract}
Image segmentation is a critical step in computational biomedical image analysis, typically evaluated using metrics like the Dice coefficient during training and validation. However, in clinical settings without manual annotations, assessing segmentation quality becomes challenging, and models lacking reliability indicators face adoption barriers. To address this gap, we propose a novel framework for predicting segmentation quality  without requiring ground truth annotations during test time. Our approach introduces two complementary frameworks: one leveraging predicted segmentation and uncertainty maps, and another integrating the original input image, uncertainty maps, and predicted segmentation maps. We present Bayesian adaptations of two benchmark segmentation models—SwinUNet and Feature Pyramid Network with ResNet50—using Monte Carlo Dropout, Ensemble, and Test Time Augmentation to quantify uncertainty. We evaluate four uncertainty estimates—confidence map, entropy, mutual information, and expected pairwise Kullback-Leibler divergence—on 2D skin lesion and 3D liver segmentation datasets, analyzing their correlation with segmentation quality metrics. Our framework achieves an R² score of 93.25 and Pearson correlation of 96.58 on the HAM10000 dataset, outperforming previous segmentation quality assessment methods. For 3D liver segmentation, Test Time Augmentation with entropy achieves an R² score of 85.03 and a Pearson correlation of 65.02, demonstrating cross-modality robustness. Additionally, we propose an aggregation strategy that combines multiple uncertainty estimates into a single score per image, offering a more robust and comprehensive assessment of segmentation quality compared to evaluating each measure independently. The proposed uncertainty-aware segmentation quality prediction network is interpreted using gradient-based methods such as Grad-CAM and feature embedding analysis through UMAP. These techniques provide insights into the model's behavior and reliability, helping to assess the impact of incorporating uncertainty into the segmentation quality prediction pipeline. The code is available at: \url{https://github.com/sikha2552/Uncertainty-Aware-Segmentation-Quality-Prediction-Bayesian-Modeling-with-Comprehensive-Evaluation-}
\end{abstract}




\end{frontmatter}


\section{Introduction}\label{sec1}

Image segmentation plays a crucial role in medical image analysis, enabling the accurate localization of organs, tumors, and lesions across various imaging modalities, including Magnetic Resonance Imaging (MRI), Computed Tomography (CT), and Ultrasound. Several methods have been proposed for segmenting regions of interest, such as region-growing \cite{nock2004statistical}, clustering \cite{lei2019automatic}, and Deep Convolutional Neural Network (DCNN)-based models \cite{valindria2017reverse}. Among these, DCNN-based models have achieved state-of-the-art accuracy due to their ability to extract complex features and patterns from training data. Despite their effectiveness in applications such as abdominal cancer analysis \cite{wang2019abdominal}, brain tumor analysis \cite{somasundaram2019current} etc., the adoption of DCNN-based segmentation models in clinical practice remains limited. A key reason for this is that most state-of-the-art segmentation models provide results without offering a measure of confidence in their predictions \cite{lin2022novel}. Evaluating segmentation quality is critical in clinical workflows, allowing clinicians to prioritize complex cases characterized by lower confidence scores from the segmentation model. Traditional approaches to assessing segmentation quality, such as boundary-based, region-based, and hybrid metrics, heavily depend on the availability of ground truth data \cite{ge2007new}, \cite{movahedi2010design}, \cite{huttenlocher1993comparing}. In the absence of ground truth for individual samples, the applicability of these metrics is limited.
The evaluation of segmentation models without ground truth is a relatively unexplored but important area, particularly in medical image analysis, where obtaining ground truth for large datasets is often impractical. Recent advances in deep learning have introduced models capable of predicting segmentation quality in the absence of ground truth  \cite{robinson2018subject}, \cite{ng2022estimating}. These models leverage features and patterns from training data to assess the quality of unlabeled images. For instance, Robinson et al. \cite{robinson2018subject} proposed a CNN-based regression model to predict Dice Similarity Coefficients (DSC) using original images and segmentation maps, while Zhou et al. \cite{zhou2019robust} combined deep reconstruction networks with regression models for DSC prediction. However, these approaches do not incorporate uncertainty measures, which are crucial for understanding model confidence and improving segmentation reliability \cite{abdar2021review}.

To address this gap, recent studies have integrated uncertainty quantification into segmentation quality assessment. DeVries et al.\cite{devries2018leveraging} introduced a DCNN-based model that uses uncertainty maps alongside original images and segmentation maps to predict DSC scores and flag cases for expert review. Similarly, Ng et al.\cite{ng2022estimating} framed quality assessment as a classification problem, leveraging uncertainty quantification for out-of-distribution detection and quality control in cardiac MRI segmentation. In our prior work \cite{sikha2024uncertainty} , we developed a framework using uncertainty estimates to predict segmentation quality without manual annotations, demonstrating its effectiveness on a skin lesion dataset. However, most existing models are limited to a single modality and rely on a single uncertainty quantification method. No studies have systematically evaluated and benchmarked multiple uncertainty models (such as Monte Carlo Dropout, Ensembles, and Test Time Augmentation) alongside various uncertainty estimates (including confidence maps, entropy, mutual information, and expected pairwise Kullback-Leibler divergence) in a multimodal setting.

Building upon these studies, this paper proposes an uncertainty-aware segmentation quality prediction framework that functions in the absence of ground truth during test time. We evaluate several uncertainty quantification methods(UMs), including: 1) Monte Carlo Dropout (MCD), 2) Ensemble models, and 3) Test Time Augmentation (TTA), to asses their effectiveness in predicting segmentation quality in combination with various uncertainty estimates(UEs), including: 1) confidence maps, 2) entropy 3) mutual information, 4) Expected pairwise Kullback-Leibler (KL) divergence. The objective is to assess how well these methods correlate with traditional image-level segmentation quality metrics such as the Dice Similarity Coefficient (DSC).  This research aims to enhance the quality and reliability of DCNN-based medical image segmentation models through advanced uncertainty quantification techniques. The major contributions of this work include:
\begin{itemize}
\item We present Bayesian versions of two benchmark semantic segmentation models: SwinUNet and FPN with ResNet50, employing three Bayesian modeling approaches—Monte Carlo Dropout (MCD), Ensemble, and Test Time Augmentation (TTA). While not explicitly Bayesian, TTA can be viewed as a Bayesian mixture model for uncertainty estimation.
\item We propose two distinct prediction architectures for segmentation quality assessment: (i) a two-branch model processing segmentation and uncertainty maps, and (ii) a three-branch model incorporating the original image, segmentation map, and uncertainty map.
\item We interpret the proposed uncertainty-aware segmentation quality prediction frameworks using state-of-the-art gradient-based methods, such as Grad-CAM, and feature embedding analysis with UMAP.
\item We provide a comprehensive evaluation of the proposed segmentation quality prediction frameworks on both 2D skin lesion and 3D liver segmentation datasets. This includes analyzing the effects of various uncertainty methods (MCD, TTA, and ensemble techniques) in combination with different uncertainty estimation strategies (Confidence maps, entropy, mutual information and KL- divergence)  and their correlation with segmentation quality metrics.
\item We propose an aggregation strategy that combines multiple uncertainty estimates into a single score per image, offering a more robust and comprehensive assessment of segmentation quality compared to evaluating each uncertainty measure independently.
      \end{itemize}

\section{Related Work}
 This section reviews the relevant literature, categorized into two main themes:  Addressing predicting segmentation quality in the absence of ground truth during test time and uncertainty computation in deep learning models.

\textbf{Segmentation quality prediction in the absence of ground truth:}
The scarcity of labeled data presents a significant challenge in training and evaluating deep learning models, particularly in medical imaging tasks such as segmentation and classification, where expert annotations are expensive and time-consuming. This limitation hinders the development of supervised models that rely on labeled datasets for training and validation. In clinical applications, insufficient labeled data can lead to unreliable model performance, increasing the risk of misdiagnoses or improper treatment.
To address this issue, Konyushkova et al. \cite{konyushkova2017learning} explored active learning to optimize the use of limited labeled data. Their approach predicts the expected error reduction for selected samples, improving model efficiency with fewer annotations. However, this technique still requires human intervention for iterative labeling.
Recent studies have proposed methods for evaluating segmentation models without ground truth. Reverse classification accuracy (RCA) \cite{valindria2017reverse} treats segmentation outputs as pseudo-ground truth labels to train a new segmentation model, using the classifier's accuracy as a proxy for segmentation quality. However, RCA depends on annotated reference datasets from similar domains, limiting its applicability in diverse clinical settings. Kohlberger et al. \cite{kohlberger2012evaluating} introduced a regression-based approach that predicts segmentation errors using hand-crafted features extracted from predicted segmentation masks. While effective, this method requires multiple annotated datasets and relies on fully supervised learning. More recent efforts have focused on integrating uncertainty measures into deep learning-based segmentation pipelines. De Vries et al. \cite{devries2018leveraging} proposed a two-stage architecture that incorporates spatial uncertainty as an intermediate representation for segmentation quality prediction. Similarly, Roy et al. \cite{roy2019bayesian} introduced a Bayesian approach to estimate segmentation quality without ground truth labels. By leveraging Bayesian networks, their method provides uncertainty quantification, enabling evaluation in data-scarce scenarios. However, its implementation is computationally expensive and complex. A recent study by Chen et al. \cite{chen2023evaluation} proposed an objective evaluation method for cell segmentation based on predefined characteristics of effective segmentation. Their approach calculates a segmentation quality score using metrics derived from the similarity between different segmentation methods, incorporating assessments with added noise and down-sampling. While this method does not require reference segmentations, its quality score is based on simple statistical features, which may not fully capture segmentation accuracy in complex medical imaging tasks.
Despite these advancements, further research is needed to develop robust benchmarking methodologies for segmentation quality prediction without ground truth. Addressing computational challenges and improving generalizability across different medical imaging domains remain key areas for future work.
\begin{figure*}
     \centering
     \includegraphics[width=7 in]{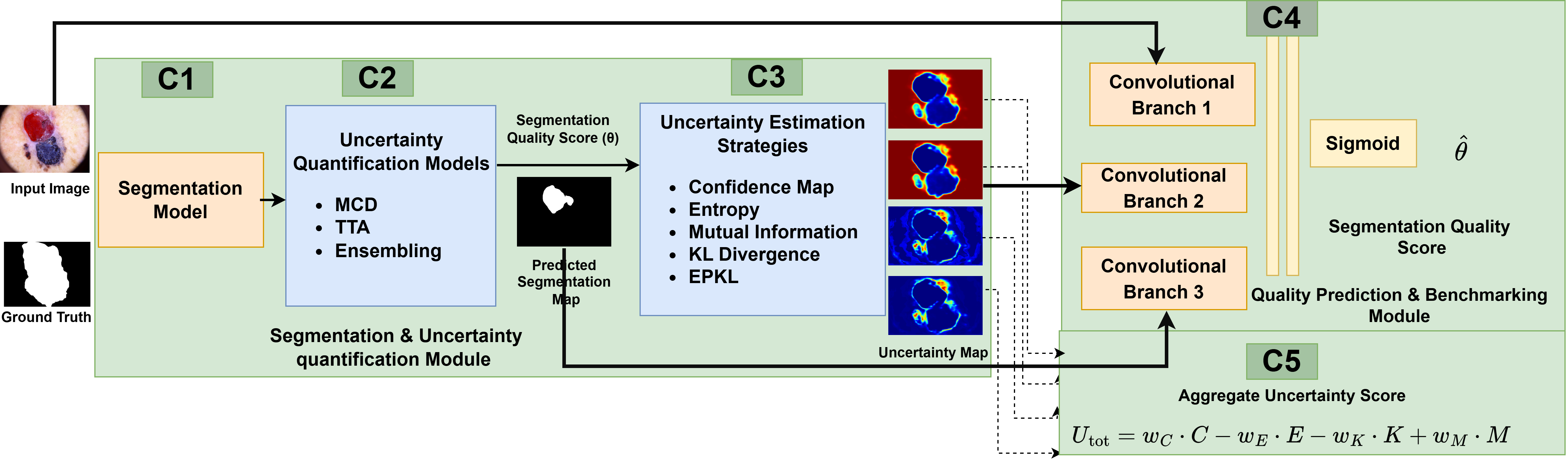}
     \caption{Overview of the Proposed Framework: Components of Uncertainty-Aware Segmentation Quality Prediction and Benchmarking.Panel 1 illustrates the segmentation and uncertainty quantification module, where C1 represents the segmentation backbone, C2 corresponds to the uncertainty models, and C3 denotes the uncertainty estimates. Panel 2 consists of the C4 quality prediction and segmentation module, while C5 represents the aggregate uncertainty score module.  }
     \label{fig:Figure1}
 \end{figure*}
\\\textbf{Uncertainty Computation and Bayesian Methods:}Understanding and measuring uncertainty in deep learning models is essential for ensuring their reliability, especially in critical fields like medical imaging. Uncertainty estimates help flag situations where a model might make a wrong prediction or encounter unfamiliar data, which is vital for real-world applications. Traditional approaches, such as Bayesian Neural Networks (BNNs) \cite{neal2012bayesian}, use Bayesian statistics to model uncertainty by treating network weights as probability distributions. This allows the model to output a range of predictions with associated confidence levels. However, BNNs are computationally impractical for large-scale use due to their complexity. Recent advancements, including Monte Carlo Dropout \cite{gal2016dropout}, Multiplicative Normalizing Flows \cite{louizos2017multiplicative}, and Stochastic Batch Normalization \cite{atanov2019uncertainty}, have improved uncertainty estimation with varying success. For instance, Wang et al. \cite{wang2019aleatoric} enhanced Bayesian methods by combining Monte Carlo Dropout with test-time augmentation (TTA), generating multiple augmented inputs to refine uncertainty estimates and boost accuracy. Despite these innovations, most techniques rely on repeated sampling—requiring 10 to 100 network evaluations during inference—leading to significant computational overhead. Deep Ensembles \cite{lakshminarayanan2017simple}, which aggregate predictions from multiple independently trained models, mitigate this to some extent but still demand substantial resources for training and inference. While these methods hold promise for robust uncertainty quantification, their adoption in medical imaging is hindered by practical barriers, underscoring the need for efficient, scalable solutions tailored to clinical work flows.
\section{Method}\label{sec3}
To effectively discuss the ability of uncertainty measures for the segmentation quality prediction, we begin by defining the key components inspired from \cite{kahl2024values} of the proposed uncertainty-aware segmentation quality prediction model as illustrated in Figure.\ref{fig:Figure1}. The key components of the model include:\\

    \begin{itemize}
    \item  \textbf{C1: Segmentation Backbone}
    \\The segmentation backbone serves as the fundamental building block of the entire architecture and remains unchanged throughout the study. The segmentation model is initially trained on the input images and the corresponding segmentation ground truth. Once trained, this model is kept fixed during the subsequent uncertainty analysis and quality prediction phases. We used SwinUNet \cite{cao2022swin} and Feature Pyramid Network (FPN) with Resnet50 \cite{lin2017feature} as the segmentation backbones.  While various segmentation models exist in the literature, we prioritized architectures that align better with our goal of improving segmentation quality through uncertainty modeling. SwinUNet’s hierarchical transformer architecture has demonstrated state-of-the-art performance in liver segmentation tasks, particularly in capturing fine-grained boundaries and heterogeneous textures in abdominal Computed Tomography(C)T scans \cite{tang2022self},\cite{cao2022swin}. Similarly, FPN with ResNet50 has been widely used for skin cancer segmentation, with Esteva et al. \cite{esteva2017dermatologist} and Codella et al. \cite{codella2019skin} highlighting its effectiveness in capturing fine-grained details in dermoscopic images. The segmentation model (S) takes the pair (x, y) as inputs and generates  ($\hat{Y}$) as the output, representing the predicted segmentation map as in equation.\ref{eq1}.
    \begin{equation}
        \label{eq1}
        (\hat{Y}) = S(x, y)
    \end{equation}
    The uncertainty estimates $U$ , derived from the segmentation prediction $\hat{Y}$, provide pixel-level measures of uncertainty, with $U$ representing an uncertainty map. In the proposed study, $U$ is calculated from the output of $S$ using different uncertainty models.%%  including Monte Carlo Dropout (MCD), Test Time Augmentation (TTA), and ensemble modeling. The final segmentation predictions are generated by taking the average of prediction maps obtained from various uncertainty models.

     \item  \textbf{C2: Uncertainty Models (UMs)}
     \\The Uncertainty Models (UMs) operate on the trained segmentation backbone and generate segmentation samples using Bayesian approximation.  In full Bayesian learning, the posterior predictive distribution is formulated as:
    \begin{equation}
        \label{eq2}
        p(y|x,D) = \int p(y|x,\theta)p(\theta|D) d\theta
    \end{equation}
    Where $D$ is the training dataset, $x$ is the input image, $y$ is the output class label, $\theta$ is the weights learned by the deep learning model, $p(y|x,\theta)$ is the likelihood and $p(\theta|D)$ is the posterior distribution. However, in practice, performing full Bayesian inference in deep learning models is computationally infeasible due to the complexity of calculating the posterior distribution. As a result, this study employs several Bayesian approximation techniques from the literature as uncertainty models to effectively estimate uncertainty without requiring full Bayesian learning, including: 1) Monte Carlo Dropout (MCD, \cite{gal2016dropout}) , which uses dropout during inference to simulate Bayesian learning, 2) Ensemble models \cite{liu2019accurate}, which combine the predictions of multiple models trained independently to capture model uncertainty)and, 3) Test Time Augmentation (TTA, \cite{wang2019aleatoric}), which generates multiple augmented versions of the input image and aggregates their predictions to estimate uncertainty. These methods allow to approximate Bayesian inference and assess the uncertainty in the segmentation predictions.
    \begin{enumerate}
        \item \textbf{Monte Carlo Dropout (MCD):}
         Monte Carlo Dropout (MCD) estimate model uncertainty in neural networks \cite{gal2016dropout} by enabling dropout during test time. In general, during training, dropout layers deactivate a subset of neurons with a probability $p$. Let $S_W(x)$ represent the output of the network for input $x$ with parameters $W$. The dropout process can be represented as $S_{\hat{W}}(x)$, where $\hat{W}$ denotes the result of element-wise multiplication between W and $\delta$. Here, $\delta$  is a binary mask drawn from a Bernoulli distribution with probability $p$, determining which elements of $W$ are retained or dropped. MCD enables dropout during inference time, and multiple forward passes are performed with different dropout masks, yielding a distribution of predictions $S_{\hat{W}}^{(t)}(x)$. This distribution provides uncertainty estimates, aiding in identifying regions where the model's predictions may be unreliable. The final segmentation map for an input image $x$, $\hat{Y}$ is produced by averaging the predictions generated for different passes.
        \begin{equation}
            \label{eq3}
            \hat{Y} = \frac{1}{T} \sum_{t=1}^{T} S_{\hat{W}}^{(t)}(x)
        \end{equation}
        Where $T$ is the number of passes, and $T=10$ in our experiment, and $\hat{Y}^{(t)}$ is the individual segmentation maps generated for each pass $t$.
         \item{\textbf{Ensemble models:}}
         Ensemble models involve combining predictions from multiple individual models to improve overall performance\cite{liu2019accurate}. In the context of uncertainty estimation, ensemble methods create diversity among models, allowing them to capture uncertainty and produce more robust predictions.
         %Given $M$ models with predictions $\hat{Y}_m$ for input $x$, where $m$ indexes models, and the ensemble prediction $\hat{Y}$ is typically computed as the average or weighted average as in Equation.\ref{eq4},
       % \begin{equation}
       %     \label{eq4}
       %     \hat{Y} = \frac{1}{M} \sum_{m=1}^{M} \hat{Y}_m
       % \end{equation}
        In our experiment, we used an ensemble of 3 models.
        \item{\textbf{Test Time Augmentation (TTA):}}
        TTA involves applying data augmentation transformations, such as rotations, flips, or color adjustments, to the input image during test time and generating predictions for each augmented version \cite{wang2019aleatoric}. The final prediction is then obtained by aggregating the predictions from all augmented versions, typically using averaging or voting. In our experiment we used 10 random rotations in test time.
    \end{enumerate}

    \item  \textbf{C3: Uncertainty Estimates (UEs)}
    \\ The uncertainty estimates (UEs) involve computing an uncertainty score for each pixel based on the predicted class scores, which can be represented as an uncertainty heat map. We used various metrics to estimate pixel-level uncertainty, including maximum confidence, entropy, mutual information, and expected pairwise Kullback-Leibler (EPKL) divergence.
    \begin{enumerate}
        \item \textbf{Confidence Map: }
        The confidence map is calculated as the maximum value across all classes over average candidate segmentation. Let $N$ represent the number of segmentation maps generated by the uncertainty model explained in the previous section. For each pixel $(i,j)$ in the input image and each class $c$, the model generates $N$ segmentation maps where $n=1,2,...,N$. The segmentation maps are averaged to produce an average probability map:
        \begin{equation}
            \label{eq5}
            \hat{P}(i,j,c) = \frac{1}{N} \sum_{n=1}^{N} P_n(i,j,c)
        \end{equation}
        The confidence map $C(i,j)$ is then calculated by taking the maximum value across all classes $c$ for each pixel $(i,j)$.
        \begin{equation}
            \label{eq6}
            C(i,j) = \max_c \hat{P}(i,j,c)
        \end{equation}
        The resulting confidence $C(i,j)$ map highlights the highest probability class for each pixel, providing a measure of the model's confidence in its predictions across the different candidate segmentation.
        \item \textbf{Entropy: }
        Renyi expected entropy is a generalization of Shannon entropy defined as:
        \begin{equation}
            \label{eq7}
            H_{\alpha}(P) = \frac{1}{1-\alpha} \log \sum_{c=1}^{C} P(c)^{\alpha}
        \end{equation}
        Where $C$ is the number of classes, and $\alpha$ controls the sensitivity of the measure; following \cite{molchanova2023novel}, we use $\alpha=2$ in our experiment.
        \item \textbf{Mutual Information (MI):}
        Mutual Information (MI) captures the disagreement between predictions. It is calculated as the difference between the entropy of the expected outcome (the entropy of the average of multiple candidate segmentation) and the expected entropy (the average of the entropy associated with each individual candidate segmentation). Let $P(c|x)$ represent the predictive distribution and $P(c|x, \theta^{(t)})$  the distribution given model parameters.% represent the predictive distribution given the model parameters $\theta^{(t)}$ for each candidate segmentation. The entropy of the expected outcome $H[P(c|x)]$ is computed as:
        The Mutual Information (MI) is computed as the difference between the entropy of the expected outcome and the expected entropy over model parameters:

     \begin{align}
     MI &= -\sum_{c=1}^C P(c \mid x) \log P(c \mid x) \notag \\
     &\quad + \frac{1}{T} \sum_{t=1}^T \sum_{c=1}^C P\left(c \mid x, \theta^{(t)}\right)
     \log P\left(c \mid x, \theta^{(t)}\right)
     \end{align}

       % \begin{equation}
        %    \label{eq8}
        %    H[P(c|x)] = - \sum_{c=1}^{C} P(c|x) \log P(c|x)
       % \end{equation}
       % The expected entropy $E_{\theta}[H[P(c|x, \theta)]]$ is computed as:
       % \begin{equation}
        %    \label{eq9}
       %     E_{\theta}[H[P(c|x, \theta)]] = - \frac{1}{T} \sum_{t=1}^{T} \sum_{c=1}^{C} P(c|x, \theta^{(t)}) \log P(c|x, \theta^{(t)})
        %\end{equation}
       % Finally, the Mutual Information is calculated as the difference between these two quantities:
        %\begin{equation}
       %     \label{eq10}
        %    MI = H[P(c|x)] - E_{\theta}[H[P(c|x, \theta)]]
        %\end{equation}
        \item \textbf{Expected pairwise Kullback-Leibler (KL) divergence (EPKL):}
        EPKL quantifies the average difference between the predicted class probabilities from individual models and the average predicted probabilities. It reflects the variability and uncertainty in the predictions. The EPKL is defined as:
        %quantifies the average difference between the predicted class probabilities across different models and the expected class probabilities. Specifically, it measures how much the predictions from individual models deviate from the average prediction. This provides an indication of the variability and uncertainty among the models' predictions. Let $P_m(c|x)$ be the predicted class probabilities from model $m$ for input $x$, and let $\hat{P}(c|x)$ be the average predicted class probabilities across all models. The EPKL divergence is calculated as follows:
        \begin{equation}
            \label{eq11}
            EPKL = \frac{1}{M} \sum_{m=1}^{M} KL(P_m(c|x) \parallel \hat{P}(c|x))
        \end{equation}
        where $P_m(c \mid x)$ is the predicted class probability from model $m$, and $\hat{P}(c \mid x)$ is the average prediction across all models. Higher EPKL values indicate greater disagreement and uncertainty among the models' predictions.
        %$KL(P \parallel Q)$ is the Kullback-Leibler divergence between the predicted probabilities of model $m$.% and the average predicted probabilities calculated as:
        %     \begin{equation}
        %     \label{eq12}
       %  KL(P \parallel Q) = \sum_{c=1}^{C} P(c) \log \frac{P(c)}{Q(c)}
       %  \end{equation}
       % This measure provides insights into the consistency of predictions across different models, with higher EPKL values indicating greater disagreement and uncertainty.
    \end{enumerate}

    \item{\textbf{C4: Segmentation Quality Prediction Model}}
    \\ The segmentation quality prediction model in our study is a regressor, specifically designed as a multi-input Convolutional Neural Network (CNN). To thoroughly analyze and benchmark the most effective input combinations for segmentation quality prediction, we developed and tested the model with two different configurations: 1) a two-way CNN,  and 2) a three-way CNN.  The two-way CNN uses two distinct input streams, allowing us to evaluate the impact of combining different types of information, such as uncertainty maps (U) and predicted segmentation maps ($\hat{Y}$) or uncertainty maps (U) and original images (Img).
%\begin{figure*}
%     \centering
%     \includegraphics[width=6in]{fig2_1.png}
%     \caption{
%Proposed frameworks for uncertainty-aware segmentation quality prediction: (a) Two-way framework using segmentation and uncertainty maps as input, and (b) Three-way framework incorporating segmentation maps, uncertainty maps, and the original input image. Both predict the Dice coefficient without access to ground truth during test time.}
   %  \label{fig:Figure2}
% \end{figure*}
  %  \begin{equation}
  %    \label{eq13}
  %        \hat{\alpha}=Q(U,(\operatorname{Img} \text { or } \hat{Y}), \alpha)
   %   \end{equation}
The three-way CNN incorporates three inputs for training: original image ($Img$), uncertainty map ($U$), predicted segmentation map ($\hat{Y}$) and is trained in a supervised manner on the segmentation quality metric.% $\alpha$ as in Equation. \ref{eq14}.
The current study uses the Dice coefficient as the segmentation quality metric  to train the prediction model. The prediction network is trained using the same training set used for training the underlying segmentation backbone. The ground truth quality metric was obtained by comparing ground truth from the dataset and segmentation predictions generated by $S$. Section.4 describes the proposed segmentation quality prediction frameworks in detail.

   %   \begin{equation}
   %   \label{eq14}
  %        \hat{\alpha}=Q(\operatorname{Img}, U, \hat{Y}, \alpha)
  %    \end{equation}

    %The current study uses the Dice coefficient as the segmentation quality metric  to train the prediction model  $Q$. The prediction network is trained using the same training set used for training the underlying segmentation backbone. The ground truth quality metric $\alpha$ was obtained by comparing ground truth from the dataset and segmentation predictions generated by $S$. Section.4 describes the proposed segmentation quality prediction frameworks in detail.

    \item{\textbf{C5: Aggregate Uncertainty Score}}
    \\The aggregation strategy combines four types of uncertainty estimates discussed before—confidence maps (C), entropy(E), EPKL, and mutual information(MI)—into a single score for each image to provide a comprehensive assessment of segmentation quality as in Equation.\ref{eq15}.

    \begin{equation}
        U_{\text {tot }}=w_C * C-w_E * E+w_M * M I-w_K * EPKL
        \label{eq15}
    \end{equation}

Each uncertainty measure is first normalized to ensure the values are in a comparable range. Weights are then assigned to each measure based on their relative importance. In our experiments, confidence maps consistently showed strong correlation with the prediction task across different models and datasets. Therefore, we assign higher weight to confidence map compared to the other measures as 0.4 , while entropy, EPKL, and mutual information are each assigned a weight of 0.2. The normalized measures are then multiplied by their respective weights, and these weighted measures are summed to produce a single aggregate score for each image. This aggregate score effectively integrates different aspects of uncertainty, offering a more robust evaluation of the segmentation performance compared to considering each measure in isolation.
    \par The images from the datasets are then sorted based on the proposed aggregate score, to visually analyze the obtained results. In most cases, this approach successfully identified images with ambiguous labeling, often characterized by large areas of incorrectly labeled pixels. This may be due to the lesion being indistinguishable from the background, leading to disagreements among annotators. Therefore,  this method could be used to flag such images for a double-check by the annotator.
\end{itemize}
\section{Uncertainty-aware segmentation quality prediction}
This section describes the two proposed uncertainty-aware segmentation quality prediction models: one consisting of two subnetworks and the other incorporating three subnetworks.
\subsection{Multi-Input CNN with Two Sub-networks for Segmentation Quality Prediction}
To construct CNN networks for predicting the segmentation quality, we initially explored two-way multi-input CNNs utilizing two distinct input combinations: 1) uncertainty estimates and predicted segmentation maps, 2) uncertainty estimates and original images.  In both scenarios, the two sub-networks  follow a consistent structure comprising five layers: 64, 64, 32, 32, and 16 filters, utilizing 3x3 filters for feature extraction and employing rectified linear unit (ReLU) activation. Subsequent to each convolutional layer, MaxPool2D layers with a 2x2 pool size are incorporated to downsample feature maps, thereby enhancing computational efficiency and controlling overfitting. Feature maps from the two branches (conv1, conv2) are concatenated, flattened, and processed through two 128-unit dense layers (ReLU) before a linear-activated output layer generates continuous quality scores (e.g., Dice score). For the first input combination, uncertainty maps and predicted segmentation maps were used. The uncertainty maps highlight low-confidence regions in the model’s predictions, pinpointing areas where segmentation reliability may deviate from the ground truth. These maps complement the predicted segmentations (the model’s best output), enabling the CNN to correlate uncertainty patterns with segmentation accuracy. This complementary information allows the network to infer quality metrics like Dice scores without ground truth. In the second configuration, uncertainty maps were paired with original images. While uncertainty maps identify regions of low confidence, the raw images provide spatial and contextual details. Together, they enable the CNN to contextualize uncertainty within the input’s inherent features, improving its ability to estimate segmentation quality. Both dual-input strategies enhance segmentation benchmarking by leveraging complementary data sources, even in ground truth-absent scenarios.

\subsection{Multi-Input CNN with Three Sub-networks for Segmentation Quality Prediction}
To provide a comprehensive feature representation for the regression task, this experiment incorporates three types of input into the regression module: the original image, uncertainty maps, and predicted segmentation maps. This three-fold input strategy is designed to leverage the strengths of each data type.  The original image offers the raw, unprocessed spatial and contextual information necessary for accurate analysis. The uncertainty maps highlight regions where the model's predictions are less confident, providing crucial insights into potential areas of error. The predicted segmentation maps, on the other hand, represent the model's best attempt at delineating different regions within the image.
\section{Datasets and Segmentation Backbones}
The study assessed the effectiveness of the proposed framework on two datasets: 1) the HAM10000 dataset for skin lesions and 2) the liver dataset, utilizing two different segmentation backbones: the Feature Pyramid Network (FPN) for the HAM10000 dataset and the Swin UNET for the liver dataset. The choice of these two networks was motivated by the distinct characteristics of the datasets and the corresponding requirements for effective segmentation \cite{lin2017feature},\cite{cao2022swin}.

The HAM10000 dataset \cite{tschandl2018ham10000} contains 2500 dermatoscopic images of pigmented skin lesions. 50\% of the images were used for training the segmentation model, while the remaining 50\% were reserved for testing. Sample images and corresponding ground truths are shown in Figure.3. The semantic segmentation network for the skin lesion segmentation  was trained using a Feature Pyramid Network with a Resnet50 backbone \cite{lin2017feature}. Feature Pyramid Networks are designed to enhance the ability of convolutional neural networks to detect objects at multiple scales by constructing a multi-scale feature pyramid from a single-resolution input image. The ResNet50 backbone processes the input image through its various convolutional layers, producing a series of feature maps at different resolutions. These feature maps capture hierarchical information, with deeper layers encoding finer details  and shallower layers retaining more abstract features. The FPN then constructs a feature pyramid by combining these multi-scale feature maps through lateral connections and top-down pathways. Each layer in the pyramid integrates information from both higher and lower resolution layers, ensuring that features at all scales are represented. The segmentation model was  trained for 5 epochs and the quality prediction network for 20 epochs, using a batch size of 16 and an Adam optimizer with a learning rate of 0.001.
 \begin{figure} [h]
     \centering
     \includegraphics[width=0.8\linewidth]{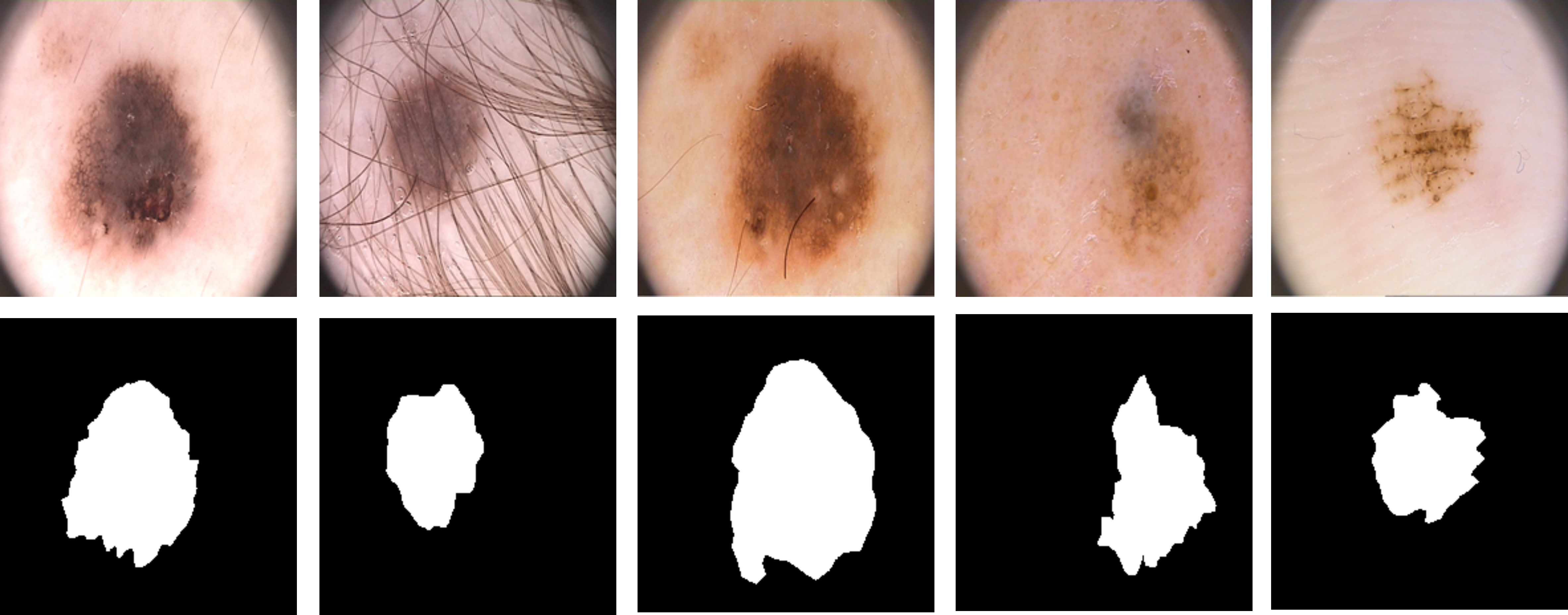}
     \caption{Sample images and respective annotations from the HAM10000 dataset \cite{tschandl2018ham10000}}
     \label{fig:Figure3}
 \end{figure}
The liver dataset used in this study was sourced from the Kaggle platform \cite{bilic2019liver} and consists of 130 CT scans for segmenting both the liver and tumor lesions. For this study, the tumor labels were merged with the liver labels to simplify the problem into a binary classification task. This dataset is part of the Liver Tumor Segmentation Challenge (LiTS17), organized in conjunction with ISBI 2017 and MICCAI 2017. The data and segmentations were provided by multiple clinical sites globally, ensuring diversity and robustness in the dataset. Figure.\ref{fig:Figure4} shows sample images from the liver dataset. The Swin UNET \cite{cao2022swin} model was employed for liver segmentation. This advanced semantic segmentation framework combines the strengths of the Swin Transformer and the UNet architecture to address the challenges of 3D medical image segmentation. The Swin Transformer backbone processes the input volumes through hierarchical stages, effectively capturing long-range dependencies and multi-scale contextual information with its shifted window approach. This ensures that both global context and fine details are incorporated in the segmentation process. This is integrated with UNet’s ability to maintain spatial localization through skip connections, which link corresponding encoder and decoder layers. The decoder then reconstructs high-resolution segmentation maps by combining the detailed spatial information from these skip connections with the hierarchical features from the encoder. This synergy between global context captured by the Swin Transformer and the precise spatial details from UNet enables SwinUnet to achieve superior segmentation performance on complex volumetric medical datasets, effectively balancing broad contextual understanding with fine-grained local delineation.
\section{Results and Discussions}
All experiments were evaluated both quantitatively and qualitatively on the two datasets mentioned in the previous section.
%The segmentation backbone models (C1) were assessed using standard semantic segmentation metrics: Accuracy (ACC), mean Accuracy (mAcc), and mean Intersection over Union (mIOU). Additionally,
The segmentation quality prediction models were evaluated based on the R² score, Pearson Correlation Coefficient (PCC) and Root Mean Squared Error (RMSE), each of which is defined below.
\\\textbf{ R-squared ($R^2$) score:} also known as the coefficient of determination. It measures how well the predicted values approximate the actual values. It is defined as:
\begin{equation}
R^2=1-\frac{\sum_{i=1}^N\left(y_i-\hat{y}_i\right)^2}{\sum_{i=1}^N\left(y_i-\bar{y}\right)^2}
\end{equation}
where: $y_i$ are the true values,$\hat{y}_i$ are the predicted values, $\bar{y}$ is the mean of the true values, and $N$ is the number of samples.
\\\textbf{ Pearson Correlation Coefficient (PCC):} quantifies the strength and direction of the linear relationship between the predicted and actual Dice scores. The PCC value ranges from -1 to 1, where a value of 1 denotes a perfect positive linear relationship, -1 signifies a perfect negative linear relationship, and 0 indicates no linear relationship. In the context of segmentation quality prediction, a high PCC between the predicted and actual Dice scores suggests a strong linear correlation, indicating that the model's predictions are closely aligned with the true values, which is defined as:
\begin{equation}
PCC=\frac{\sum_{i=1}^N\left(y_i-\bar{y}\right)\left(\hat{y}_i-\bar{y}\right)}{\sqrt{\sum_{i=1}^N\left(y_i-\bar{y}\right)^2 \sum_{i=1}^N\left(\hat{y}_i-\bar{y}\right)^2}}
\end{equation}
where, $y_i$ and $\hat{y}_i$ are the individual values of the predicted and actual Dice scores, $\bar{x}$ and $\bar{y}$ are the mean values of the predicted and actual Dice scores, and $N$ is the number of data points.
\\ \textbf{Root Mean Squared Error (RMSE): }Measures the difference between predicted and actual Dice score values by calculating the square root of the mean squared differences. Since it penalizes larger errors more heavily due to squaring, RMSE is particularly sensitive to outliers. It is defined as:
\begin{equation}
R M S E=\sqrt{\frac{1}{N} \sum_{i=1}^N\left(y_i-\hat{y}_i\right)^2}
\end{equation}
where, N is the total number of samples, $y_i$ represents the true and $\hat{y}_i$ denotes the predicted Dice scores.
 \begin{figure} [tb]
    \centering
     \includegraphics[width=1\linewidth]{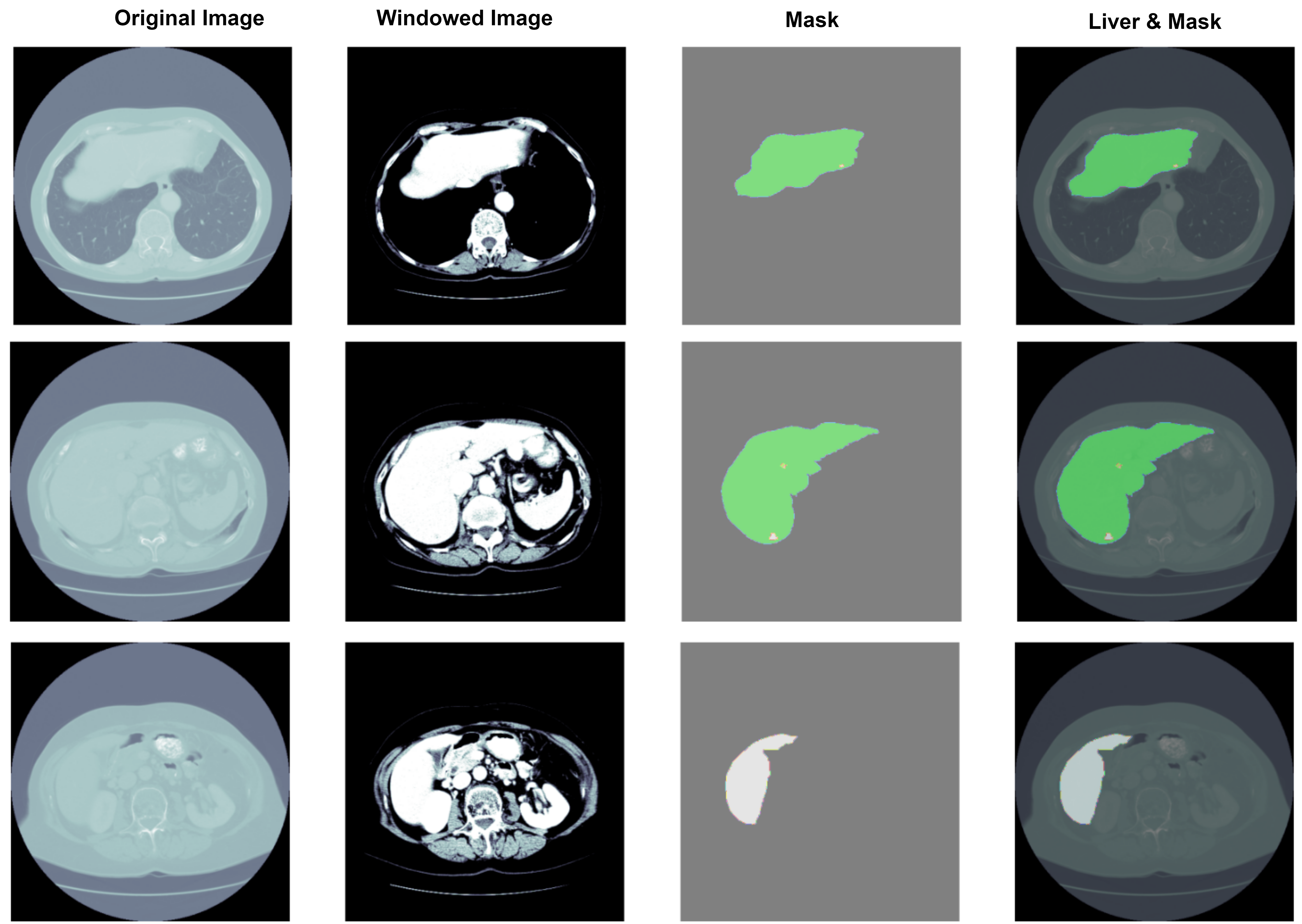}
     \caption{Sample images and respective annotations from the Liver dataset \cite{tschandl2018ham10000} }
     \label{fig:Figure4}
 \end{figure}
\subsection{Quantitative analysis based on R² score and Pearson Correlation Coefficient (PCC)}
We analyzed combinations of three Uncertainty Models (UMs) for the prediction task in two-way and three-way input scenarios. The UMs included: ensemble (5 models), Monte Carlo Dropout (10 samples during test time), and Test Time Augmentation (10 random rotations during test time). The UMs were paired with four Uncertainty Estimates (UEs): confidence map, mutual information, entropy, and Expected Pairwise KL-Divergence. %This comprehensive evaluation framework allowed us to benchmark the performance and reliability of the models in predicting segmentation quality.
\begin{table*}[]
\centering
\caption{Analysis of the Effectiveness of Uncertainty-Aware Segmentation Quality Prediction  for  Multi-Input CNN with Two Sub-networks Using UE and Predicted Segmentation maps. The best results for each UM are marked in bold.}
\label{tab:1}
\begin{tabular}{|l|cc|cc|}
\hline
\textbf{Datasets}              & \multicolumn{2}{c|}{\textbf{HAM10000}}                                                                                                           & \multicolumn{2}{c|}{\textbf{Liver}}                                                                                         \\ \hline
\textbf{UM+UE}                 & \multicolumn{1}{c|}{\textbf{R2 Score}} & \multicolumn{1}{l|}{\textbf{\begin{tabular}[c]{@{}l@{}}Pearson Correlation\\ Coefficient\end{tabular}}} & \multicolumn{1}{c|}{\textbf{R2 Score}} & \textbf{\begin{tabular}[c]{@{}c@{}}Pearson Correlation\\ Coefficient\end{tabular}} \\ \hline
\textbf{Ensemble + Confidence} & \multicolumn{1}{c|}{\textbf{86.07}}    & \textbf{92.83}                                                                                          & \multicolumn{1}{c|}{\textbf{72.94}}    & \textbf{85.88}                                                                     \\ \hline
Ensemble + MI                  & \multicolumn{1}{c|}{81.86}             & 90.53                                                                                                   & \multicolumn{1}{c|}{66.88}             & 84.35                                                                              \\ \hline
Ensemble + Epkl                & \multicolumn{1}{c|}{80.71}             & 89.98                                                                                                   & \multicolumn{1}{c|}{71.13}             & 85.08                                                                              \\ \hline
Ensemble + Entropy   & \multicolumn{1}{c|}{85.19}    & 92.34                                                                                          & \multicolumn{1}{c|}{70.1}              & 84.37                                                                              \\ \hline
\textbf{MCD + Confidence}      & \multicolumn{1}{c|}{\textbf{93.25}}    & \textbf{96.58}                                                                                          & \multicolumn{1}{c|}{\textbf{61.63}}    & \textbf{78.81}                                                                     \\ \hline
MCD + MI                       & \multicolumn{1}{c|}{91.13}             & 95.8                                                                                                    & \multicolumn{1}{c|}{60.81}             & 77.98                                                                              \\ \hline
MCD + Epkl                     & \multicolumn{1}{c|}{92.64}             & 96.6                                                                                                    & \multicolumn{1}{c|}{57.91}             & 77.41                                                                              \\ \hline
MCD + Entropy                  & \multicolumn{1}{c|}{90.61}             & 95.21                                                                                                   & \multicolumn{1}{c|}{60.61}             & 78.11                                                                              \\ \hline
\textbf{TTA + Confidence}      & \multicolumn{1}{c|}{\textbf{71.7}}     & \textbf{85.74}                                                                                          & \multicolumn{1}{c|}{\textbf{71.6}}     & \textbf{85.4}                                                                      \\ \hline
TTA + MI                       & \multicolumn{1}{c|}{66.6}              & 82.81                                                                                                   & \multicolumn{1}{c|}{70.8}              & 84.7                                                                               \\ \hline
TTA + Epkl                     & \multicolumn{1}{c|}{61.79}             & 79.95                                                                                                   & \multicolumn{1}{c|}{70.0}              & 83.3                                                                               \\ \hline
TTA + Entropy                  & \multicolumn{1}{c|}{71.17}             & 84.74                                                                                                   & \multicolumn{1}{c|}{70.4}              & 83.4                                                                               \\ \hline
\end{tabular}
\end{table*}
Table.\ref{tab:1} presents the effectiveness of uncertainty-aware segmentation quality prediction for the Multi-Input CNN with two sub-networks (UE and Predicted Segmentation maps as inputs) on  the HAM10000 and Liver datasets. %The model is evaluated on the HAM10000 and Liver datasets using different Uncertainty Models (UMs) combined with different Uncertainty Estimates (UEs) discussed earlier.
Model predictions were measured in terms of the R² Score and the Pearson Correlation Coefficient (PCC). The best results for each UM are highlighted.  For the HAM10000 dataset, ensemble models combined with confidence maps yielded an R² score of 86.07 and a PCC of 92.83, while Monte Carlo Dropout (MCD) with confidence map achieved the highest performance with an R² score of 93.25 and a PCC of 96.58. Test Time Augmentation (TTA) with confidence map showed comparatively lower performance, with an R² score of 71.7 and a PCC of 85.74. On the Liver dataset, ensemble models with confidence map produced an R² score of 72.94 and a PCC of 85.88, while MCD with confidence map had an R² score of 61.63 and a Pearson correlation of 78.81. Similar to the HAM10000 dataset, TTA combined with all uncertainty estimations (UEs) showed comparatively lower performance on the Liver dataset.
\begin{table*}[]
\centering
\caption{Analysis of the Effectiveness of Uncertainty-Aware Segmentation Quality Prediction  for  Multi-Input CNN with Two Sub-networks Using UE and original images. Best results for each UM is marked in bold.}
\label{tab:2}
\begin{tabular}{|l|cc|cc|}
\hline
\textbf{Datasets} & \multicolumn{2}{c|}{\textbf{HAM10000}} & \multicolumn{2}{c|}{\textbf{Liver}} \\ \hline
\textbf{UM+UE} & \multicolumn{1}{c|}{\textbf{R2 Score}} & \multicolumn{1}{l|}{\textbf{\begin{tabular}[c]{@{}l@{}}Pearson Correlation \\ Coefficient\end{tabular}}} & \multicolumn{1}{c|}{\textbf{R2 Score}} & \textbf{\begin{tabular}[c]{@{}c@{}}Pearson Correlation\\  Coefficient\end{tabular}} \\ \hline
\textbf{Ensemble + Confidence} & \multicolumn{1}{c|}{\textbf{69.35}} & \textbf{83.47} & \multicolumn{1}{c|}{\textbf{63.72}} & \textbf{56.39} \\ \hline
Ensemble + MI & \multicolumn{1}{c|}{66.3} & 81.51 & \multicolumn{1}{c|}{63.34} & 53.67 \\ \hline
Ensemble + Epkl & \multicolumn{1}{c|}{64.84} & 80.52 & \multicolumn{1}{c|}{62.43} & 51.81 \\ \hline
Ensemble + Entropy & \multicolumn{1}{c|}{67.64} & 82.33 & \multicolumn{1}{c|}{61.38} & 50.78 \\ \hline
\textbf{MCD + Confidence} & \multicolumn{1}{c|}{\textbf{57.99}} & \textbf{76.65} & \multicolumn{1}{c|}{\textbf{61.48}} & \textbf{53.53} \\ \hline
MCD + MI & \multicolumn{1}{c|}{54.51} & 74.73 & \multicolumn{1}{c|}{62.34} & 52.31 \\ \hline
MCD + Epkl & \multicolumn{1}{c|}{55.63} & 74.62 & \multicolumn{1}{c|}{61.43} & 51.95 \\ \hline
MCD + Entropy & \multicolumn{1}{c|}{48.3} & 73.08 & \multicolumn{1}{c|}{60.79} & 51.72 \\ \hline
TTA + Confidence & \multicolumn{1}{c|}{66.13} & 82.99 & \multicolumn{1}{c|}{63.45} & 57.71 \\ \hline
TTA + MI & \multicolumn{1}{c|}{63.47} & 80.19 & \multicolumn{1}{c|}{61.53} & 53.32 \\ \hline
TTA + Epkl & \multicolumn{1}{c|}{61.14} & 79.01 & \multicolumn{1}{c|}{63.55} & 56.01 \\ \hline
\textbf{TTA + Entropy} & \multicolumn{1}{c|}{\textbf{71.12}} & \textbf{84.71} & \multicolumn{1}{c|}{\textbf{63.75}} & \textbf{56.50} \\ \hline
\end{tabular}
\end{table*}
\begin{figure*}
     \centering
     \includegraphics[width=1\linewidth]{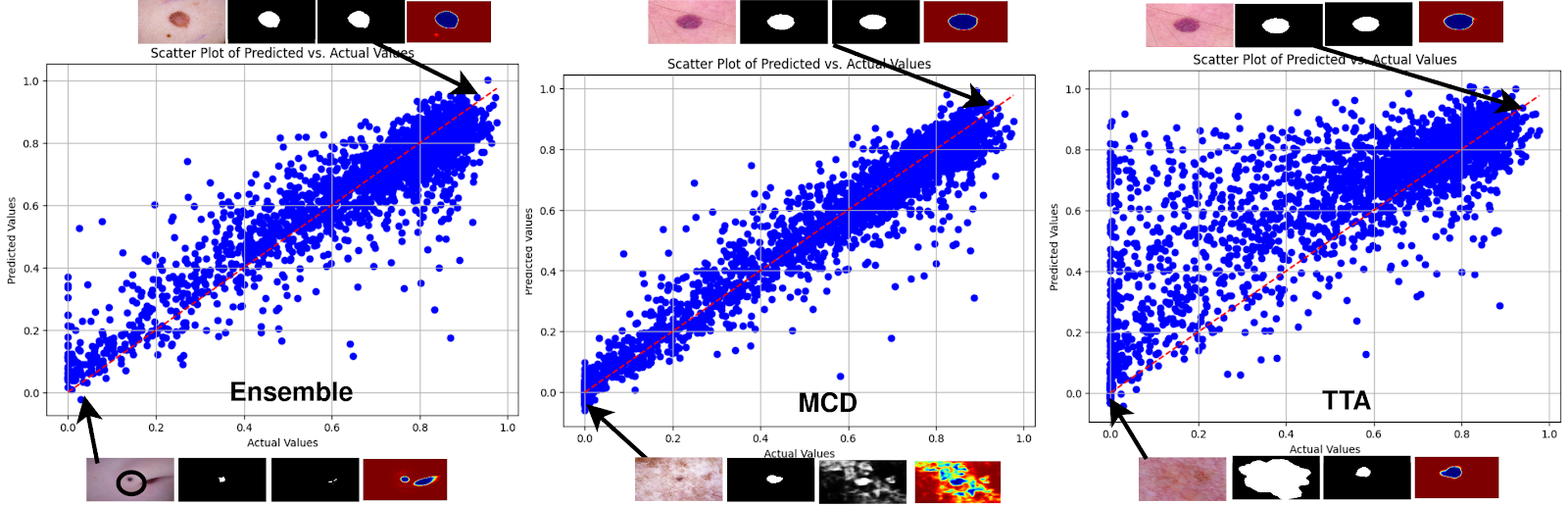}
     \caption{Scatter plot of predicted Dice score and original Dice score from the proposed quality prediction framework, using confidence as UE, and Ensembling, MCD, and TTA as UMs. Sample images associated with some points are shown (Original image, Ground truth, Prediction, Confidence map). Confidence maps are able to capture false positives, please see black circle.}
    \label{fig:Figure5}
 \end{figure*}
The results for the two-way CNN suggest that confidence maps consistently provide better performance across all UMs. Confidence maps directly reflect the highest predicted probability for each pixel, making them a robust indicator of prediction reliability. This property likely contributes to their superior performance compared to other UEs, as the model can effectively identify and leverage the most confident predictions. Among UMs, ensemble models consistently performed well across both datasets, particularly when combined with confidence maps. The strength of ensemble models lies in their ability to reduce variance and increase robustness by aggregating the predictions of multiple models. The ensemble approach resolve the risk of over-fitting to specific artifacts or noise present in individual model predictions, effectively addressing epistemic uncertainty. Monte Carlo Dropout also demonstrated strong performance, particularly on the HAM10000 dataset, where MCD combined with confidence maps achieved the highest R² score of 93.25 and a PCC of 96.58. This method simulates a variety of model configurations, capturing the uncertainty in the model's parameters , thereby addressing epistemic uncertainty.
%On the Liver dataset, the combination of MCD and Confidence had an R² score of 61.63 and a PCC of 78.81. The high performance of MCD with confidence maps underscores their  ability to capture epistemic uncertainty to improve segmentation quality predictions.
Test Time Augmentation (TTA), which primarily addresses aleatoric uncertainty by introducing variability in input data, showed lower performance in predicting segmentation quality. While it performed reasonably well on the HAM10000 dataset, with an R² score of 71.7 and a PCC of 85.74 using confidence maps, its overall performance was lower. The reduced effectiveness may be due to noise or irrelevant variations introduced by augmentations, suggesting that TTA's effectiveness is highly dependent on the dataset and augmentation type.

 Table.\ref{tab:2} presents the performance of two-way CNN using uncertainty estimates (UE) and original images as inputs. The results, evaluated on the HAM10000 and Liver datasets, show that TTA + Entropy achieved the highest R² score (71.12) and Pearson correlation (84.71) on HAM10000, while Ensemble + Confidence delivered the best R² score (63.72) on the Liver dataset. In conclusion, for the two-way input CNN models, combining predicted segmentation maps with uncertainty estimates demonstrated stronger correlation with segmentation quality metrics. Confidence maps consistently improved segmentation quality predictions across various UMs by clearly identifying reliable predictions. Ensemble models and MCD, which address epistemic uncertainty, significantly benefit from confidence maps, whereas TTA, capturing aleatoric uncertainty, showed variable performance due to the potential introduction of noise or artifacts.
 \begin{table*}[]
\centering
\caption{ Analysis of the Effectiveness of Uncertainty-Aware Segmentation Quality Prediction  for  Multi-Input CNN with Three Sub-networks Using UE and original images. Best results for each UM is marked in bold.}
\label{tab:3}
\begin{tabular}{|l|cc|cc|}
\hline
\textbf{Datasets} & \multicolumn{2}{c|}{\textbf{HAM10000}} & \multicolumn{2}{c|}{\textbf{Liver}} \\ \hline
\textbf{UM+UE} & \multicolumn{1}{c|}{\textbf{R2 Score}} & \multicolumn{1}{l|}{\textbf{\begin{tabular}[c]{@{}l@{}}Pearson Correlation\\ Coefficient\end{tabular}}} & \multicolumn{1}{c|}{\textbf{R2 Score}} & \textbf{\begin{tabular}[c]{@{}c@{}}Pearson Correlation \\ Coefficient\end{tabular}} \\ \hline
Ensemble + Confidence & \multicolumn{1}{c|}{84.88} & 92.28 & \multicolumn{1}{c|}{68.72} & 83.12 \\ \hline
Ensemble + MI & \multicolumn{1}{c|}{84.2} & 91.93 & \multicolumn{1}{c|}{71.06} & 85.31 \\ \hline
Ensemble + Epkl & \multicolumn{1}{c|}{80.63} & 90.54 & \multicolumn{1}{c|}{70.04} & 84.16 \\ \hline
\textbf{Ensemble + Entropy} & \multicolumn{1}{c|}{\textbf{86.49}} & \textbf{93.2} & \multicolumn{1}{c|}{\textbf{70.03}} & \textbf{84.76} \\ \hline
MCD + Confidence & \multicolumn{1}{c|}{93.44} & 96.87 & \multicolumn{1}{c|}{59.28} & 77.31 \\ \hline
MCD + MI & \multicolumn{1}{c|}{92.72} & 96.29 & \multicolumn{1}{c|}{58.52} & 77.2 \\ \hline
MCD + Epkl & \multicolumn{1}{c|}{90.32} & 95.16 & \multicolumn{1}{c|}{58.71} & 77.41 \\ \hline
\textbf{MCD + Entropy} & \multicolumn{1}{c|}{\textbf{93.99}} & \textbf{97.01} & \multicolumn{1}{c|}{\textbf{59.31}} & \textbf{77.88} \\ \hline
TTA + Confidence & \multicolumn{1}{c|}{87.8} & 94.13 & \multicolumn{1}{c|}{84.76} & 64.88 \\ \hline
TTA + MI & \multicolumn{1}{c|}{85} & 92.3 & \multicolumn{1}{c|}{84.29} & 63.51 \\ \hline
TTA + Epkl & \multicolumn{1}{c|}{72.34} & 90.28 & \multicolumn{1}{c|}{84.52} & 64.23 \\ \hline
\textbf{TTA + Entropy} & \multicolumn{1}{c|}{\textbf{89.13}} & \textbf{94.49} & \multicolumn{1}{c|}{\textbf{85.03}} & \textbf{65.02} \\ \hline
\end{tabular}
\end{table*}
Figure.\ref{fig:Figure5} compares predicted versus actual Dice scores from the two-way input CNN, using predicted segmentation maps and uncertainty estimates (UEs) as inputs. The segmentation tasks utilize confidence maps combined with uncertainty measures. Each scatter plot shows predicted Dice scores (y-axis) against actual Dice scores (x-axis), with a red dashed line representing the ideal match scenario (y=x). Sample images, including the original image, ground truth, prediction, and confidence map, are shown below each scatter plot to illustrate the results. For the Ensemble method (left plot), the points align well along the red dashed line, indicating accurate predictions. There is a notable clustering of points in the lower-left corner, showing Ensemble's effectiveness in identifying poor-quality segmentations. The confidence map highlights areas of potential false positives, as marked by black circles. The MCD method (middle plot) also shows a good alignment along the red dashed line, similar to Ensemble, and effectively identifies poor-quality segmentations, as evidenced by the distinct cluster in the lower-left corner. The confidence maps for MCD similarly capture areas of uncertainty. In contrast, the TTA method (right plot) displays a more scattered distribution of points, particularly in the lower-left corner, indicating less reliability in identifying poor-quality segmentations. The sample images associated with each method demonstrate the ability of confidence maps to highlight areas of false positives, helping to assess the segmentation quality. Overall, Ensemble and MCD show better performance in predicting Dice scores and identifying poor-quality segmentations compared to TTA.
\begin{figure}[!ht]
     \centering
     \includegraphics[height=3 in]{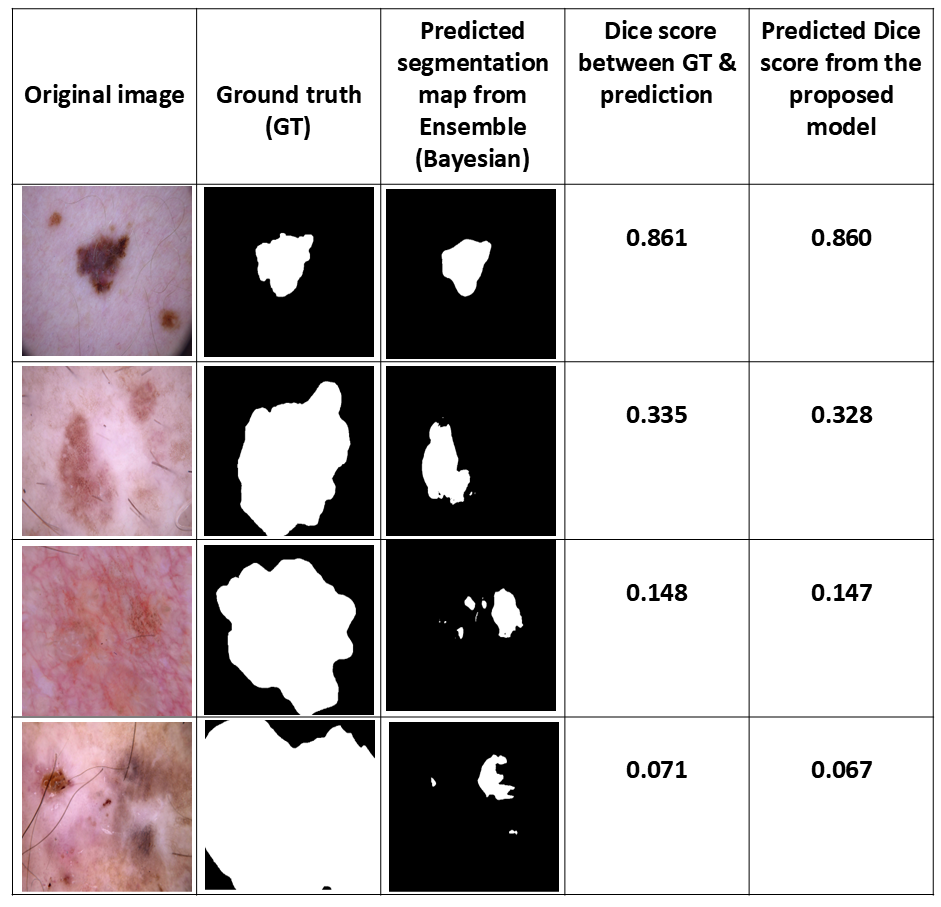}
     \caption{ Comparison of Ground Truth Segmentation, Bayesian Model Predictions, and Dice Score Estimates from the Proposed Uncertainty-Aware Segmentation Quality Prediction Model}
     \label{fig:cmpgt}
 \end{figure}
\begin{figure*}
     \centering
     \includegraphics[width=7in]{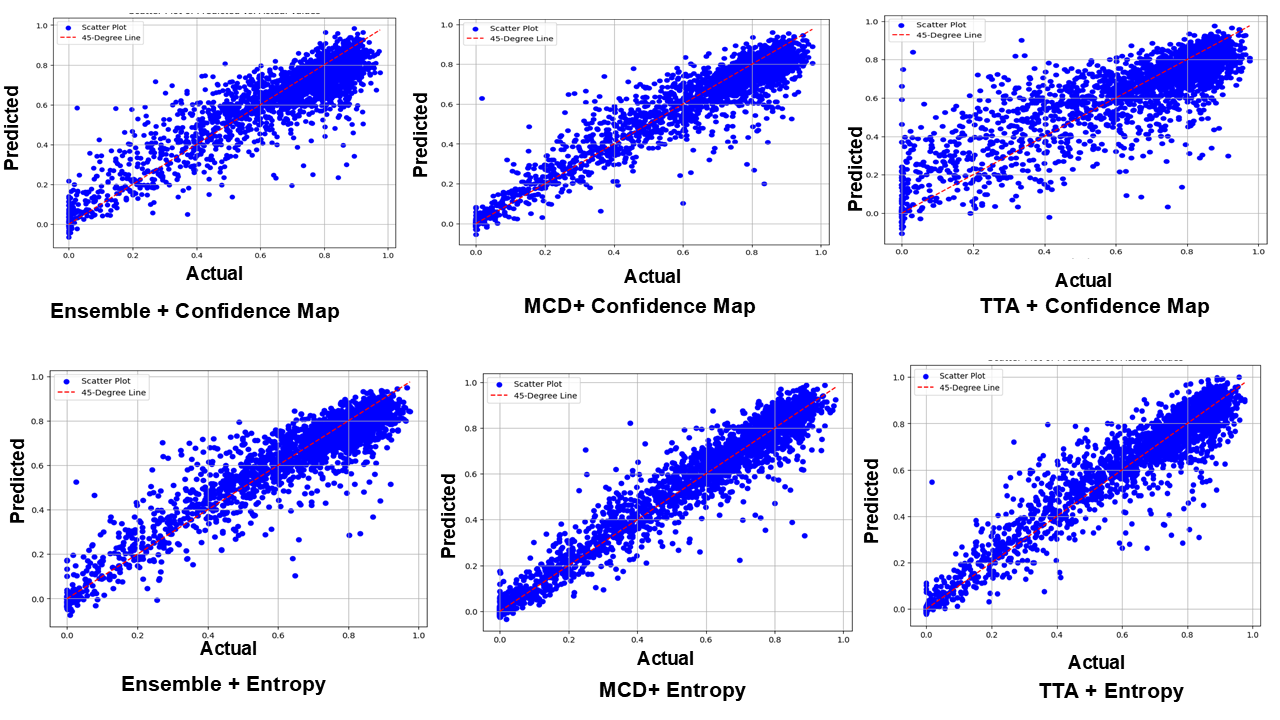}
     \caption{Scatter Plot of Predicted vs. Original Dice Scores from the Proposed Quality Prediction Frameworks on Skin Dataset. The First Row Represents Best Input Combinations for Two-Way Prediction Model (UE and Predicted
Segmentation maps), and the Second Row Represents Best Input Combinations for Three-Way Input Model.}
     \label{fig:skinreg}
 \end{figure*}
 \begin{figure*}
     \centering
     \includegraphics[width=7in]{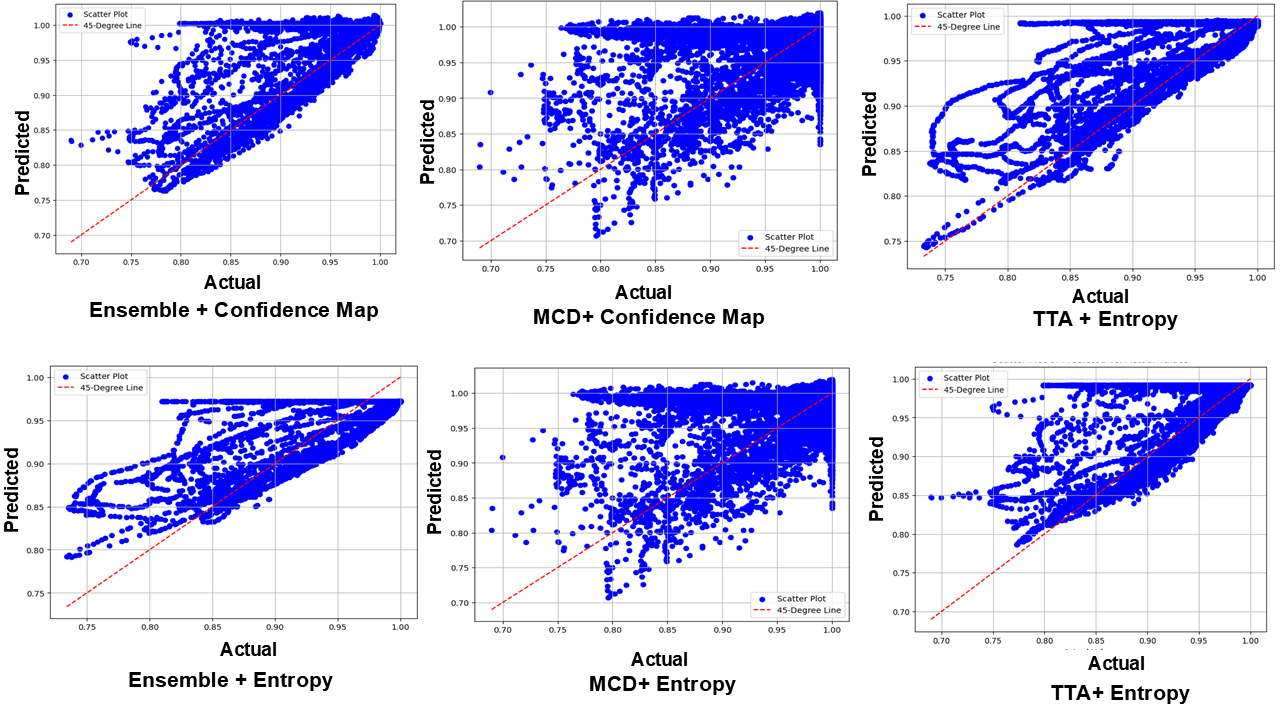}
     \caption{Scatter Plot of Predicted vs. Original Dice Scores from the Proposed Quality Prediction Frameworks on Liver Dataset. The First Row Represents Best Input Combinations for Two-Way Prediction Model(UE and Predicted
Segmentation maps), and the Second Row Represents Best Input Combinations for Three-Way Input Model.}
     \label{fig:liverreg}
 \end{figure*}
 \begin{figure}[!ht]
     \centering
     \includegraphics[height=3 in]{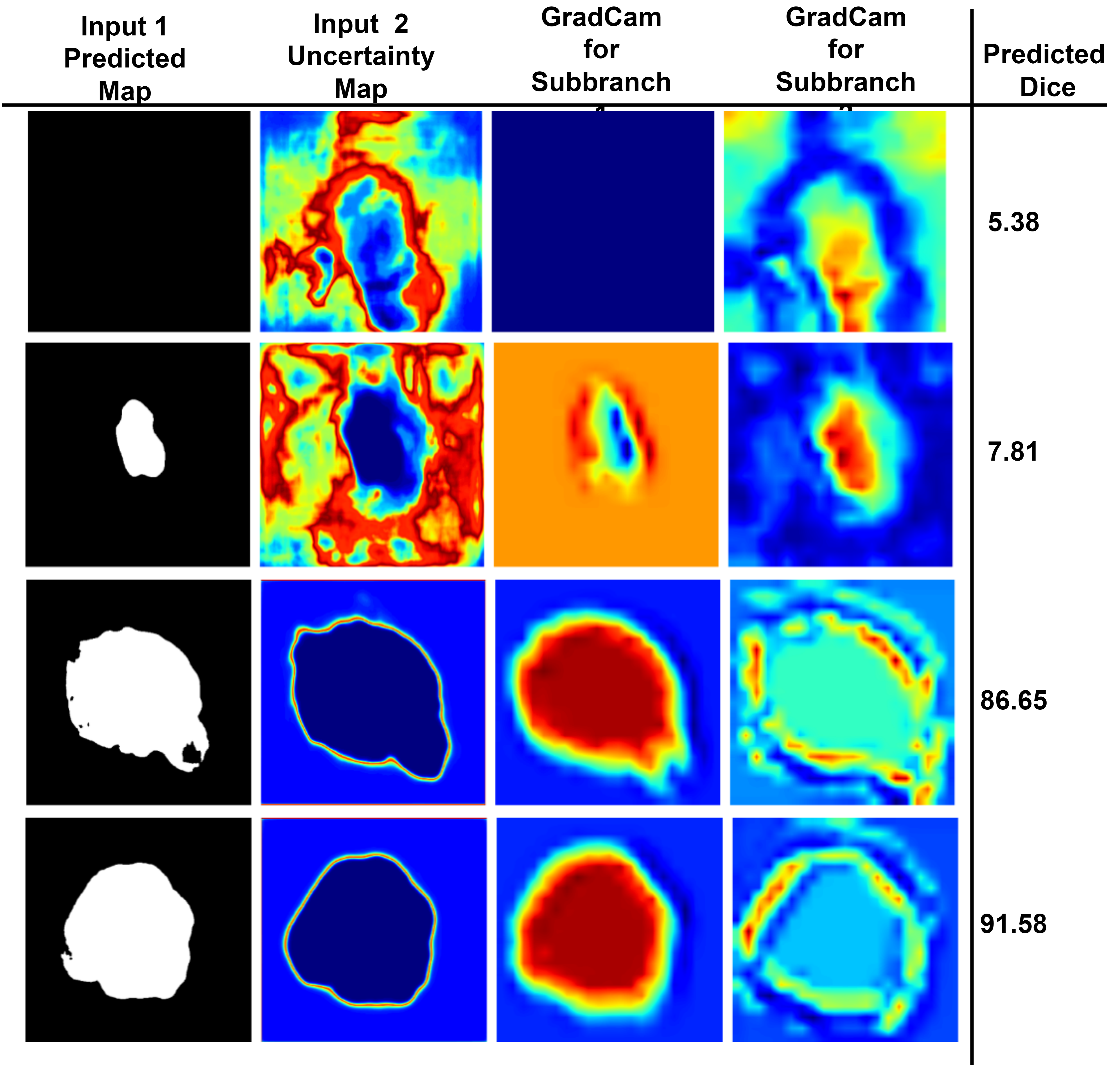}
     \caption{Grad-CAM Analysis of Segmentation Quality: Comparing Low and High Dice Score Predictions from the Two-Way segmentation quality prediction model.}
     \label{fig:Figure6}
 \end{figure}
  \begin{figure}[!ht]
    \centering
    \includegraphics[width=1\linewidth]{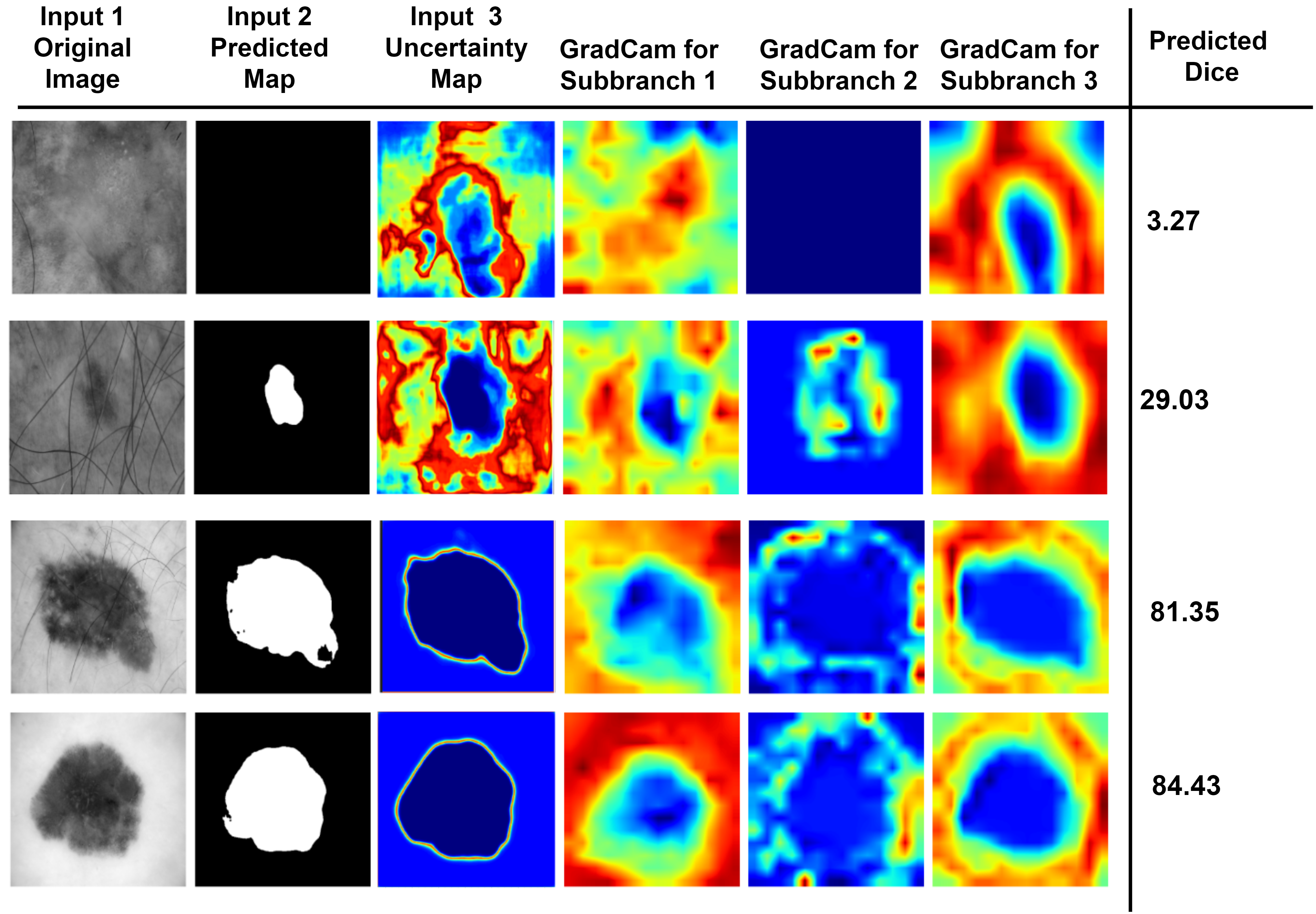}
    \caption{Grad-CAM Analysis of Segmentation Quality: Comparing Low and High Dice Score Predictions from the Three-Way segmentation quality prediction model.}
    \label{fig:Figure7}
\end{figure}

Table.\ref{tab:3} summarizes the performance of various Uncertainty Models (UMs) combined with Uncertainty Estimates (UEs) for predicting segmentation quality using a three-input setup. Unlike 2-way CNN models for quality prediction, for a 3-way CNN entropy maps showed better correlation in combination with all UMs. For the HAM10000 dataset, the Ensemble + Entropy combination achieved the highest R² score of 86.49 and a Pearson correlation of 93.2, indicating its superior ability to capture reliable segmentation quality predictions. MCD + Entropy followed closely with an R² score of 93.99 and a Pearson correlation of 97.01, demonstrating the effectiveness of Monte Carlo Dropout combined with entropy maps. TTA + Entropy also performed well, with an R² score of 89.13 and a Pearson correlation of 94.49, although it was slightly less effective compared to ensemble and MCD models. On the Liver dataset, Ensemble + MI achieved the highest R² score of 71.06 and a Pearson correlation of 85.31, suggesting that mutual information is particularly effective in this context. Compared to the two-input combinations discussed previously, the three-input setup consistently showed improved performance, highlighting the added value of incorporating original images, uncertainty maps, and predicted segmentation maps for more accurate segmentation quality prediction. Figure.\ref{fig:skinreg} and Figure.\ref{fig:liverreg} show regression scatter plots for the best input combinations of the two-way and three-way segmentation quality prediction models for the skin lesion and liver datasets, respectively.
\subsection{Validation and Comparison  with Expert Annotations and State-of-the-Art Models}
Figure.\ref{fig:cmpgt}  compares the proposed uncertainty-aware segmentation quality prediction model against expert-generated manual segmentation masks (ground truth) across four cases, demonstrating close alignment between predicted Dice scores and expert evaluations. For instance, in a high-quality segmentation case, the predicted score (0.860) matches the expert assessment (0.861), while in a low-quality case, the prediction (0.067) closely mirrors the manual evaluation (0.071). This validation against manual annotations confirms the model’s ability to approximate human judgment across a wide range of segmentation performance (Dice scores: 0.861 to 0.071). Notably, the consistency in results underscores the method’s potential for clinical deployment, offering a reliable alternative to resource-intensive manual quality checks in real-world medical settings.
We further evaluate the proposed uncertainty-aware segmentation quality prediction method against state-of-the-art models on the skin cancer segmentation dataset using RMSE as the evaluation metric. As shown in Table.\ref{tab:RMSE}, lower RMSE values indicate better performance. Ensemble-Confidence achieves the lowest RMSE (0.132 ± 0.039), demonstrating its effectiveness. Other uncertainty-based methods, such as Max Probability (0.168 ± 0.014) and LCE \cite{devries2018learning} (0.167 ± 0.019), also perform competitively. In contrast, traditional methods like RCA \cite{valindria2017reverse} exhibit higher RMSE, underscoring the advantages of uncertainty-aware approaches.

{Our findings demonstrate that explicitly incorporating uncertainty significantly improves segmentation quality prediction, reducing RMSE by up to 0.066. Among uncertainty estimation techniques, HCNN provides the least performance gain, aligning with prior research \cite{abdar2021review} that suggests aleatoric uncertainty, modeled by HCNN, is less effective in detecting model failures as it primarily captures data noise. QualityNet performs similarly to the No Uncertainty baseline, as expected due to their similar design. RCA \cite{valindria2017reverse} performs poorly, likely because it was designed for datasets with minimal inter-sample variation, such as MRI scans, whereas the HAM10000 dataset exhibits significant variability in lesion characteristics. These results highlight the benefits of uncertainty-aware approaches for segmentation quality prediction.

\begin{table}[]
\centering
\caption{Comparison of various state-of-the-art segmentation quality prediction methods on the skin cancer segmentation dataset, evaluated in terms of Root Mean Square Error (RMSE).}
\label{tab:RMSE}
\begin{tabular}{|l|l|}
\hline
\multicolumn{1}{|c|}{\textbf{Method}} & \multicolumn{1}{c|}{\textbf{RMSE ↓}} \\ \hline
RCA \cite{valindria2017reverse} & 0.438±0.007 \\ \hline
Quality Net \cite{huang2016qualitynet} & 0.213±0.009 \\ \hline
No Uncertainty & 0.198±0.011 \\ \hline
Max Probability & 0.168±0.014 \\ \hline
HCNN \cite{kendall2017uncertainties} & 0.196±0.023 \\ \hline
LCE \cite{devries2018learning} & 0.167±0.019 \\ \hline
\textbf{Ensemble- Confidence} & \textbf{0.132±0.039} \\ \hline
TTA-Confidence & 0.153±0.049 \\ \hline
MCD-Confidence & 0.209±0.071 \\ \hline
\end{tabular}
\end{table}

 \subsection{Visualization and Interpretation of Grad-CAMs and UMAPs for Segmentation Quality Assessment}
\begin{figure*}
    \centering
    \includegraphics[width=7 in]{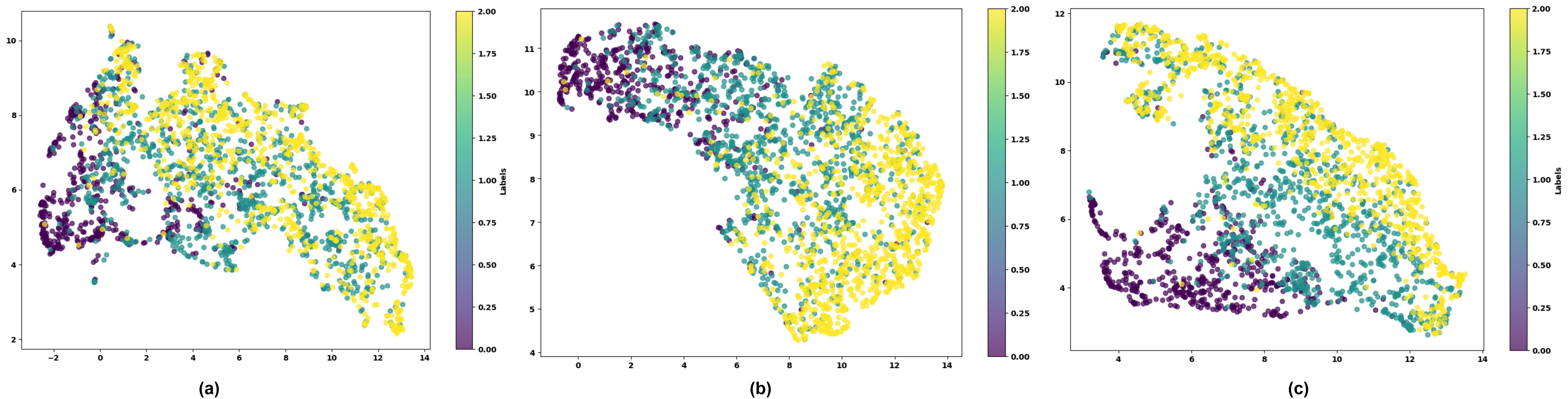}
    \caption{UMAP Embedding Visualization of Feature Separation in Two-way CNNs and  Three-way CNN for segmentation quality prediction. a) UMAP  embedding for 2-way CNN with original image and Uncertainty map as input 2) UMAP  embedding for 2-way CNN with segmentation map  and Uncertainty map as input 3) UMAP  embedding for 3-way CNN with original image, segmentation map  and Uncertainty map as input.}
    \label{fig:Figure8}
\end{figure*}
To further explain the effectiveness of confidence maps in predicting segmentation quality in a two-way set up , and entropy maps in a three-way set up, we visualize Grad-CAM  for individual subbranches of both the architectures. For calculating Grad-Cam, we considered the last convolutional layers of the sub-branches. Using gradient tape, we calculate the gradients of the predicted score with respect to the feature map activations from the convolutional layers. These gradients are averaged to obtain importance weights for each feature map. The weighted combination of these activations produces a raw heatmap, which is then normalized to values between 0 and 1 for better visualization. Figure.\ref{fig:Figure6} and Figure.\ref{fig:Figure7} show the Grad-CAMs derived from the two-way regressor and the three-way regressor, respectively. These figures illustrate the worst and best prediction samples, based on the predicted Dice score, providing insights into the model's focus during these predictions.
In Figure.\ref{fig:Figure6}, the first two rows show sample images with the lowest Dice score predictions (5.38 and 7.81, respectively) from the model, while the third and fourth rows display samples with the highest Dice score predictions (86.65 and 91.58, respectively). For samples with the lowest Dice scores, column.1 shows the predicted segmentation maps which clearly fail to accurately capture the lesion regions. The corresponding Grad-CAMs from the first sub-branch of the quality prediction model prove this by focusing more on the background, indicating the segmentation prediction is unsuccessful in identifying the lesion foreground. The 4th  column, representing the Grad-CAM from the second sub-branch, concentrates on very small regions inside the lesions, reflecting high uncertainty across most pixels as seen in the uncertainty map. This prove that when the model is uncertain, it fails to localize the lesion accurately. In contrast, for samples with the highest Dice scores, the predicted segmentation maps in the first column accurately capture the lesion regions. Corresponding Grad-CAMs from Sub-branch one (3rd  column) focuses on the predicted foreground, which corresponds well with the lesion area. Grad-CAMs from the second Sub-branch  (4th column) focuses on the edges of the lesion, aligning with the higher uncertainty at the boundaries as indicated by the uncertainty map. This dual attention mechanism allows the model to refine the segmentation by using the confidence in the foreground and the uncertainty at the edges.

Figure.\ref{fig:Figure7} visualizes the Grad-CAM for the same set of images in Figure.\ref{fig:Figure6} for the three-way regressor. From the figure it is evident that the three-input model leverages the original image for context, the uncertainty map to highlight regions of model uncertainty, and the segmentation map to define lesion boundaries, resulting in enhanced feature extraction and improved prediction accuracy. For low Dice score predictions, the Grad-CAMs from Subbranch 1 highlight irrelevant background areas, indicating the model's inability to accurately identify the lesion. The uncertainty maps confirm this with high uncertainty across most pixels. Conversely, for high Dice score predictions, the predicted segmentation maps align with the lesion regions, and the Grad-CAMs from Subbranches 1 and 3 focus on the foreground and lesion boundaries, reflecting improved confidence and segmentation precision.

To further understand the feature level separability of models, we compute the UMAP embeddings for all the three proposed models. Figure.\ref{fig:Figure8} compares the UMAP embeddings of features extracted from 2-way and 3-way CNNs to analyze the models' ability for feature separation. High-dimensional features were extracted from the final dense layer and mapped to a 2D plane for better visualization. To illustrate the feature separation capability of the learned models, predictions were grouped into three classes based on the Dice score obtained from the backbone segmentation model: Dice $<=$ 50 labeled as poor segmentation, 51 $<= \text{Dice} <= $ 80 labeled as good segmentation, and Dice $>=$ 81 labeled as best segmentation. In Figure.\ref{fig:Figure8}  column  (a), the 2-way CNN embedding with the original image and uncertainty map shows a moderate level of feature separation, indicating that these two inputs provide some ability to distinguish between different types of segmentation but with significant overlap. Column (b) represents the UMAP embedding for the 2-way CNN with the segmentation map and uncertainty map as inputs, which shows improved feature separation compared to (a), suggesting that the inclusion of the segmentation map enhances the model's ability to differentiate between better and poorer segmentations. Column (c) demonstrates the UMAP embedding for the 3-way CNN that includes the original image, segmentation map, and uncertainty map. It shows the highest degree of feature separation among the three configurations. By incorporating three modalities of information, the model effectively pushes samples apart, resulting in predicted Dice scores for bad and good predictions being significantly different. The improved clustering in (c) signifies that the 3-way CNN is better equipped to push the Dice scores further apart, thereby enhancing its predictive capabilities and robustness in segmentation tasks.
\begin{table}[]
\caption{Top 5 Images from the Liver Dataset Ranked by Aggregated Uncertainty Score and Corresponding Predicted Dice Scores.}
\label{tab:4}
\begin{tabular}{|c|cc|}
\hline
\textbf{Image ID} & \multicolumn{2}{c|}{\textbf{Predicted Dice Score}} \\ \hline
\textbf{volume-26\_slice\_67} & \multicolumn{1}{c|}{\textbf{0.68}} & \textbf{0.412} \\ \hline
\textbf{volume-26\_slice\_68} & \multicolumn{1}{c|}{\textbf{0.69}} & \textbf{0.405} \\ \hline
volume-91\_slice\_96 & \multicolumn{1}{c|}{0.71} & 0.395 \\ \hline
volume-91\_slice\_97 & \multicolumn{1}{c|}{0.72} & 0.374 \\ \hline
\textbf{Volume-70\_slice\_436} & \multicolumn{1}{c|}{\textbf{0.74}} & \textbf{0.357} \\ \hline
\end{tabular}
\end{table}
\subsection{Quantitative results based on the aggregated uncertainty score}
We performed a quantitative analysis  to assess the effectiveness of proposed aggregated uncertainty score as in equation.\ref{eq15} in ranking images based on  the presence of ambiguous or noisy labels.  Initially, we computed the aggregated uncertainty score for each image in the test set and arranged them in descending order, prioritizing those with the highest uncertainty scores first.
In Figure.\ref{fig:Figure9}, we present the top three images from this sorted list, which exhibit the highest aggregated uncertainty scores.  It is evident from the images that all three have ambiguous or noisy labels, where a significant portion of pixels are labeled as lesions,  and the model predictions are notably inaccurate. The proposed aggregated uncertainty score effectively identifies and quantifies these discrepancies between the ground truth and model predictions. This capability suggests its potential utility in flagging images for annotators to revisit and  revise their annotations, thereby enhancing the overall quality and reliability of the dataset used for training and evaluation.
\begin{figure}
    \centering
    \includegraphics[width=1\linewidth, height=5 cm]{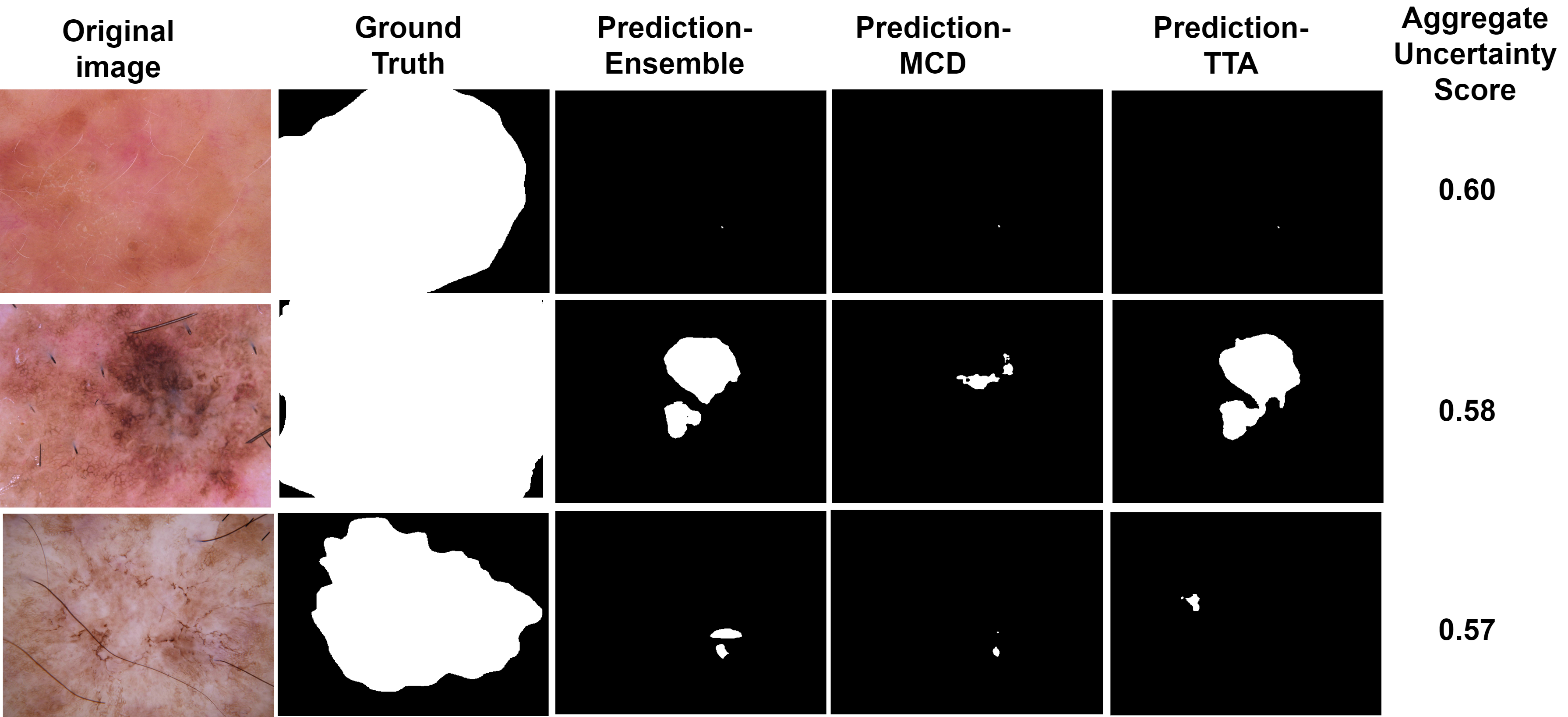}
    \caption{Top Three Images from the Test Set with Highest Aggregated Uncertainty Scores and Corresponding Model Predictions}
    \label{fig:Figure9}
\end{figure}
We evaluated the effectiveness of the proposed aggregated uncertainty score in ranking images based on the presence of ambiguous or noisy labels in the liver dataset. The results are presented in Table.\ref{tab:4}, which includes the top 5 images from the test set in terms of aggregated uncertainty score and their corresponding predicted Dice scores. The table clearly shows that the aggregated uncertainty score effectively identifies images with potential label inaccuracies, thereby helping to flag them for further review and annotation refinement, to enhance overall dataset quality and model performance.
\section{Conclusion}
This paper presents an uncertainty-aware framework for predicting segmentation quality without relying on ground truth data, focusing on the influence of various  Uncertainty Models (UM) and Uncertainty Estimating mechanisms (UE) on performance. We explore three UMs—Ensemble, Monte Carlo Dropout (MCD), and Test Time Augmentation (TTA)—in combination with four UEs— maximum confidence, entropy, mutual information, and expected pairwise KL-Divergence. These combinations are meticulously benchmarked for medical image segmentation tasks on the HAM10000 skin lesion dataset and liver CT scan segmentation dataset. Bayesian versions of two benchmark semantic segmentation models, SwinUNet for the skin lesion dataset and FPN with ResNet50 for the liver dataset, were developed. Additionally two distinct prediction architectures were proposed: the first features two sub-branches processing the predicted segmentation map and uncertainty map, while the second employs three sub-branches, incorporating the original input image along with the predicted segmentation and uncertainty maps. Comprehensive evaluation showed that Confidence maps consistently improved segmentation quality predictions across various UMs by clearly identifying reliable predictions. We used the gradient based approach-GradCAM and feature embedding approach UMAP to interpret the proposed prediction models. Furthermore, we conducted a quantitative analysis to evaluate the effectiveness of the proposed aggregated uncertainty score. By calculating this score for each image in the test set and sorting them in descending order, the results demonstrated that the aggregated uncertainty score accurately identifies and quantifies discrepancies between the ground truth and model predictions. This suggests its potential value in flagging images that require revisiting or revision by annotators, thereby improving the overall quality and reliability of the dataset used for both training and evaluation. Future work may explore further refinements to the proposed methodologies and their potential for broader applications across various imaging modalities.
\section{Conflict of Interest}
This project is co-funded by Proyectos de Colaboración PúblicoPrivada- CPP2021-008364, funded by MCIN/AEI, and the European Union through the NextGenerationEU/PRTR.

%%\appendix

%%\section{Sample Appendix Section}
%%\label{sec:sample:appendix}
%%Lorem ipsum dolor sit amet, consectetur adipiscing elit, sed do eiusmod tempor section \ref{sec:sample1} incididunt ut labore et dolore magna aliqua. Ut enim ad minim veniam, quis nostrud exercitation ullamco laboris nisi ut aliquip ex ea commodo consequat. Duis aute irure dolor in reprehenderit in voluptate velit esse cillum dolore eu fugiat nulla pariatur. Excepteur sint occaecat cupidatat non proident, sunt in culpa qui officia deserunt mollit anim id est laborum.

 \bibliographystyle{elsarticle-num}
 \bibliography{cas-refs}





% \end{thebibliography}
\end{document}